\documentclass[10pt,twocolumn,letterpaper]{article}

\usepackage{cvpr}
\usepackage{times}
\usepackage{epsfig}
\usepackage{graphicx}
\usepackage{amsmath}
\usepackage{amssymb}

\usepackage{cite}
\usepackage{ctable}
\usepackage{hhline}
\usepackage{booktabs}
\usepackage{subcaption}
\usepackage{csquotes }

\newcolumntype{L}[1]{>{\raggedright\let\newline\\\arraybackslash\hspace{0pt}}m{#1}}
\newcolumntype{C}[1]{>{\centering\let\newline\\\arraybackslash\hspace{0pt}}m{#1}}
\newcolumntype{R}[1]{>{\raggedleft\let\newline\\\arraybackslash\hspace{0pt}}m{#1}}

\usepackage[breaklinks=true,bookmarks=false]{hyperref}

\cvprfinalcopy 

\def\cvprPaperID{****} 

\ifcvprfinal\fi
\begin{document}

\title{V2V-PoseNet: Voxel-to-Voxel Prediction Network for Accurate 3D Hand and Human Pose Estimation from a Single Depth Map}

\author{Gyeongsik Moon\\
Department of ECE, ASRI\\
Seoul National University\\
{\tt\small mks0601@snu.ac.kr}
\and
Ju Yong Chang\\
Department of EI\\
Kwangwoon University\\
{\tt\small juyong.chang@gmail.com}
\and
Kyoung Mu Lee\\
Department of ECE, ASRI\\
Seoul National University\\
{\tt\small kyoungmu@snu.ac.kr}
}

\maketitle

\begin{abstract}
Most of the existing deep learning-based methods for 3D hand and human pose estimation from a single depth map are based on a common framework that takes a 2D depth map and directly regresses the 3D coordinates of keypoints, such as hand or human body joints, via 2D convolutional neural networks (CNNs). The first weakness of this approach is the presence of perspective distortion in the 2D depth map. While the depth map is intrinsically 3D data, many previous methods treat depth maps as 2D images that can distort the shape of the actual object through projection from 3D to 2D space. This compels the network to perform perspective distortion-invariant estimation. The second weakness of the conventional approach is that directly regressing 3D coordinates from a 2D image is a highly non-linear mapping, which causes difficulty in the learning procedure. To overcome these weaknesses, we firstly cast the 3D hand and human pose estimation problem from a single depth map into a voxel-to-voxel prediction that uses a 3D voxelized grid and estimates the per-voxel likelihood for each keypoint. We design our model as a 3D CNN that provides accurate estimates while running in real-time. Our system outperforms previous methods in almost all publicly available 3D hand and human pose estimation datasets and placed first in the HANDS 2017 frame-based 3D hand pose estimation challenge. The code is available in \footnote{\url{https://github.com/mks0601/V2V-PoseNet_RELEASE}}.
\end{abstract}

\section{Introduction}

Accurate 3D hand and human pose estimation is an important requirement for activity recognition with diverse applications, such as human-computer interaction or augmented reality~\cite{romero2009monocular}. It has been studied for decades in computer vision community and has attracted considerable research interest again due to the introduction of low-cost depth cameras.

\begin{figure}[t]
\begin{center}
   \includegraphics[width=1.0\linewidth]{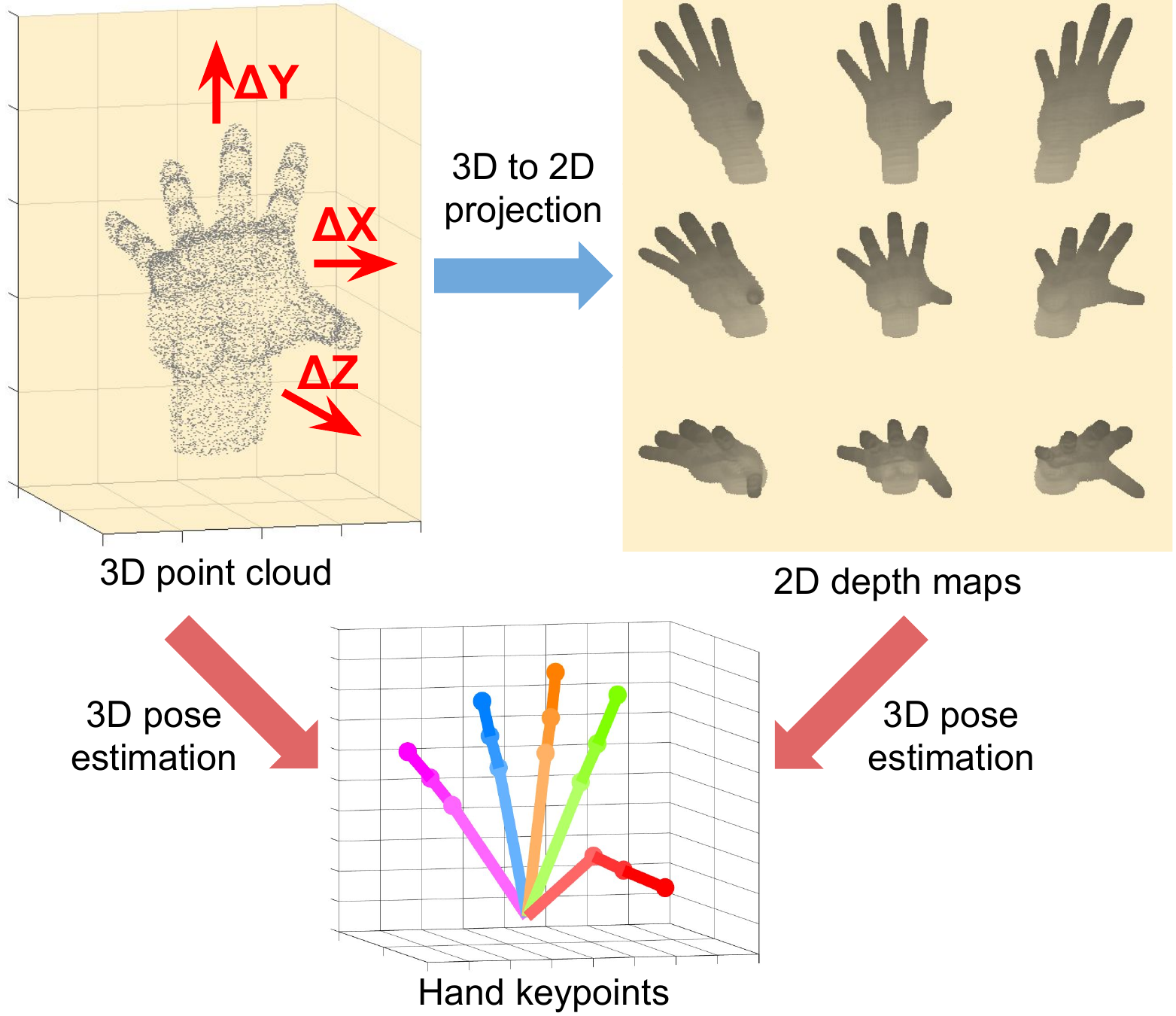}
\end{center}
\vspace*{-5mm}
   \caption{Visualization of perspective distortion in 2D depth image. The 3D point cloud has one-to-one relation with a 3D pose, but the 2D depth image has many-to-one relation because of perspective distortion. Thus, the network is compelled to perform perspective distortion-invariant estimation. The 2D depth maps are generated by translating the 3D point cloud by $\Delta$X = -300, 0, 300 mm (from left to right) and $\Delta$Y = -300, 0, 300 mm (from bottom to top). In all cases, $\Delta$Z is set to 0 mm. Similar values to the real human hand size and camera projection parameters in the MSRA dataset were used for our visualization.}
\vspace*{-3mm}
\label{fig:weakness_of_prev}
\end{figure}

Recently, powerful discriminative approaches based on convolutional neural networks (CNNs) are outperforming existing methods in various computer vision tasks including 3D hand and human pose estimation from a single depth map~\cite{ge20173d,guo2017towards,Oberweger_2017_ICCV_Workshops,chen2017pose,haque2016towards}. Although these approaches achieved significant advancement in 3D hand and human pose estimation, they still suffer from inaccurate estimation because of severe self-occlusions, highly articulated shapes of target objects, and low quality of depth images. Analyzing previous deep learning-based methods for 3D hand and human pose estimation from a single depth image, most of these methods~\cite{oberweger2015hands,oberweger2015training,bouchacourt2016disco,Wan_2017_CVPR,guo2017ren,guo2017towards,Oberweger_2017_ICCV_Workshops,chen2017pose,madadi2017end,fourure2017multi,haque2016towards} are based on a common framework that takes a 2D depth image and directly regresses the 3D coordinates of keypoints, such as hand or human body joints. However, we argue that this approach has two serious drawbacks. The first one is perspective distortion in 2D depth image. As the pixel values of a 2D depth map represent the physical distances of object points from the depth camera, the depth map is intrinsically 3D data. However, most previous methods simply take depth maps as a 2D image form, which can distort the shape of an actual object in the 3D space by projecting it to the 2D image space. Hence, the network \emph{see} a distorted object and is burdened to perform distortion-invariant estimation. We visualize the perspective distortions of the 2D depth image in Figure~\ref{fig:weakness_of_prev}. The second weakness is the highly non-linear mapping between the depth map and 3D coordinates. This highly non-linear mapping hampers the learning procedure and prevents the network from precisely estimating the coordinates of keypoints as argued by Tompson \etal~\cite{tompson2014joint}. This high nonlinearity is attributed to the fact that only one 3D coordinate for each keypoint has to be regressed from the input.

\begin{figure*}
\begin{center}
\includegraphics[width=1.0\linewidth]{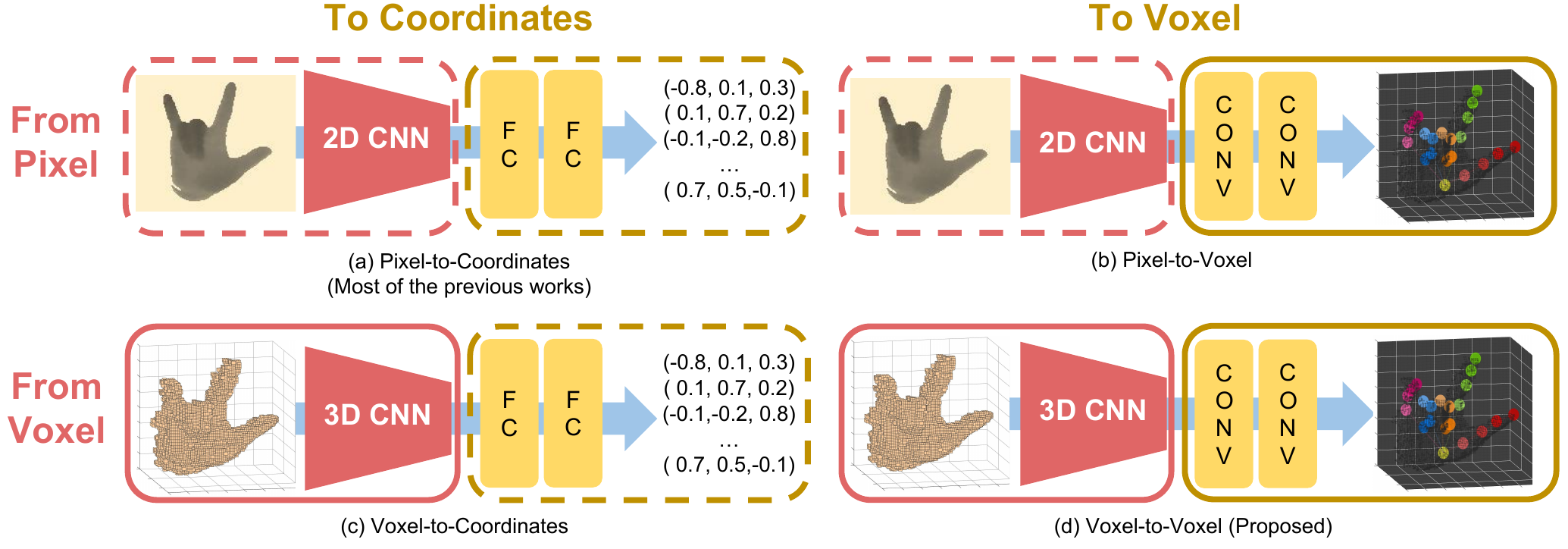}
\end{center}
\vspace*{-5mm}
   \caption{Various combinations of inputs and outputs for 3D pose estimation from a single depth image. Most of the previous works take a 2D depth image as input and estimate the 3D coordinates of keypoints as in (a). In contrast, the proposed system takes a 3D voxelized grid and estimates the per-voxel likelihood of each keypoint as in (d). Note that (b) and (d) are solely composed of the convolutional layers that become the fully convolutional architecture.}
\label{fig:comparison_io_type}
\end{figure*}

To cope with these limitations, we propose the \emph{voxel-to-voxel prediction network for pose estimation (V2V-PoseNet)}. In contrast to most of the previous methods, the V2V-PoseNet takes a voxelized grid as input and estimates the per-voxel likelihood for each keypoint as shown in Figure~\ref{fig:comparison_io_type}. 

By converting the 2D depth image into a 3D voxelized form as input, our network can \emph{sees} the actual appearance of objects without perspective distortion. Also, estimating the per-voxel likelihood of each keypoint enables the network to learn the desired task more easily than the highly non-linear mapping that estimates 3D coordinates directly from the input. We perform a thorough experiment to demonstrate the usefulness of the proposed volumetric representation of input and output in 3D hand and human pose estimation from a single depth map. The performance of the four combinations of input (i.e., 2D depth map and voxelized grid) and output (i.e., 3D coordinates and per-voxel likelihood) types are compared.

The experimental results show that the proposed voxel-to-voxel prediction allows our method to achieve the state-of-the-art performance in almost all of the publicly available datasets (i.e., three 3D hand~\cite{tompson2014real,tang2014latent,sun2015cascaded} and one 3D human~\cite{haque2016towards} pose estimation datasets) while it runs in real-time. We also placed first in the HANDS 2017 frame-based 3D hand pose estimation challenge~\cite{yuan20172017}. We hope that the proposed system to become a milestone of 3D hand and human pose estimation problems from a single depth map. Now, we assume that the term \enquote{3D pose estimation} refers to the localization of the hand or human body keypoints in 3D space.

Our contributions can be summarized as follows. 
\begin{itemize}
\item We firstly cast the problem of estimating 3D pose from a single depth map into a voxel-to-voxel prediction. Unlike most of the previous methods that regress 3D coordinates directly from the 2D depth image, our proposed V2V-PoseNet estimates the per-voxel likelihood from a voxelized grid input. 

\item We empirically validate the usefulness of the volumetric input and output representations by comparing the performance of each input type (i.e., 2D depth map and voxelized grid) and output type (i.e., 3D coordinates and per-voxel likelihood).

\item We conduct extensive experiments using almost all of the existing 3D pose estimation datasets including three 3D hand and one 3D human pose estimation datasets. We show that the proposed method produces significantly more accurate results than the state-of-the-art methods. The proposed method also placed first in the HANDS 2017 frame-based 3D hand pose estimation challenge.
\end{itemize}

\section{Related works}

{\bf Depth-based 3D hand pose estimation.} 
Hand pose estimation methods can be categorized into generative, discriminative, and hybrid methods. Generative methods assume a pre-defined hand model and fit it to the input depth image by minimizing hand-crafted cost functions ~\cite{sharp2015accurate, tang2015opening}. Particle swam optimization (PSO)~\cite{sharp2015accurate}, iterative closest point (ICP)~\cite{tagliasacchi2015robust}, and their combination~\cite{qian2014realtime} are the common algorithms used to obtain optimal hand pose results.

Discriminative methods directly localize hand joints from an input depth map. Random forest-based methods~\cite{keskin2012hand,tang2013real,tang2014latent,liang2014parsing,sun2015cascaded,tang2015opening,wan2016hand} provide fast and accurate performance. However, they utilize hand-crafted features and are overcome by recent CNN-based approaches ~\cite{tompson2014real,oberweger2015hands,ge2016robust,sinha2016deephand,bouchacourt2016disco,yang2016hand,deng2017hand3d,ge20173d,guo2017ren,guo2017towards,chen2017pose,madadi2017end,fourure2017multi,xu2017lie,Choi_2017_ICCV,Oberweger_2017_ICCV_Workshops} that can learn useful features by themselves. Tompson \etal~\cite{tompson2014real} firstly utilized CNN to localize hand keypoints by estimating 2D heatmaps for each hand joint. Ge \etal~\cite{ge2016robust} extended this method by exploiting multi-view CNN to estimate 2D heatmaps for each view. Ge \etal~\cite{ge20173d} transformed the 2D input depth map to the 3D form and estimated 3D coordinates directly via 3D CNN. Guo \etal~\cite{guo2017ren,guo2017towards} proposed a region ensemble network to accurately estimate the 3D coordinates of hand keypoints and Chen \etal~\cite{chen2017pose} improved this network by iteratively refining the estimated pose. Oberweger \etal~\cite{Oberweger_2017_ICCV_Workshops} improved their preceding work~\cite{oberweger2015hands} by utilizing recent network architecture, data augmentation, and better initial hand localization.

Hybrid methods are proposed to combine the generative and discriminative approach. Oberweger \etal~\cite{oberweger2015training} trained discriminative and generative CNNs by a feedback loop. Zhou \etal~\cite{zhou2016model} pre-defined a hand model and estimated the parameter of the model instead of regressing 3D coordinates directly. Ye \etal~\cite{ye2016spatial} used spatial attention mechanism and hierarchical PSO. Wan \etal~\cite{Wan_2017_CVPR} used two deep generative models with a shared latent space and trained discriminator to estimate the posterior of the latent pose.

{\bf Depth-based 3D human pose estimation.} 
Depth-based 3D human pose estimation methods also rely on generative and discriminative models. The generative models estimate the pose by finding the correspondences between the pre-defined body model and the input 3D point cloud. The ICP algorithm is commonly used for 3D body tracking~\cite{ganapathi2012real,grest2005nonlinear,knoop2006sensor,helten2013personalization}. Another method such as template fitting with Gaussian mixture models ~\cite{ye2014real} was also proposed. By contrast, the discriminative models do not require body templates and they directly estimate the positions of body joints. Conventional discriminative methods are mostly based on random forests. Shotton \etal~\cite{shotton2013real} classified each pixel into one of the body parts, while Girchick \etal~\cite{girshick2011efficient} and Jung \etal~\cite{jung2016sequential} directly regressed the coordinates of body joints. Jung \etal~\cite{yub2015random} used a random tree walk algorithm (RTW), which reduced the running time significantly. Recently, Haque \etal~\cite{haque2016towards} proposed the viewpoint-invariant pose estimation method using CNN and multiple rounds of a recurrent neural network. Their model learns viewpoint-invariant features, which makes the model robust to viewpoint variations.

\begin{figure*}
\begin{center}
\includegraphics[width=1.0\linewidth]{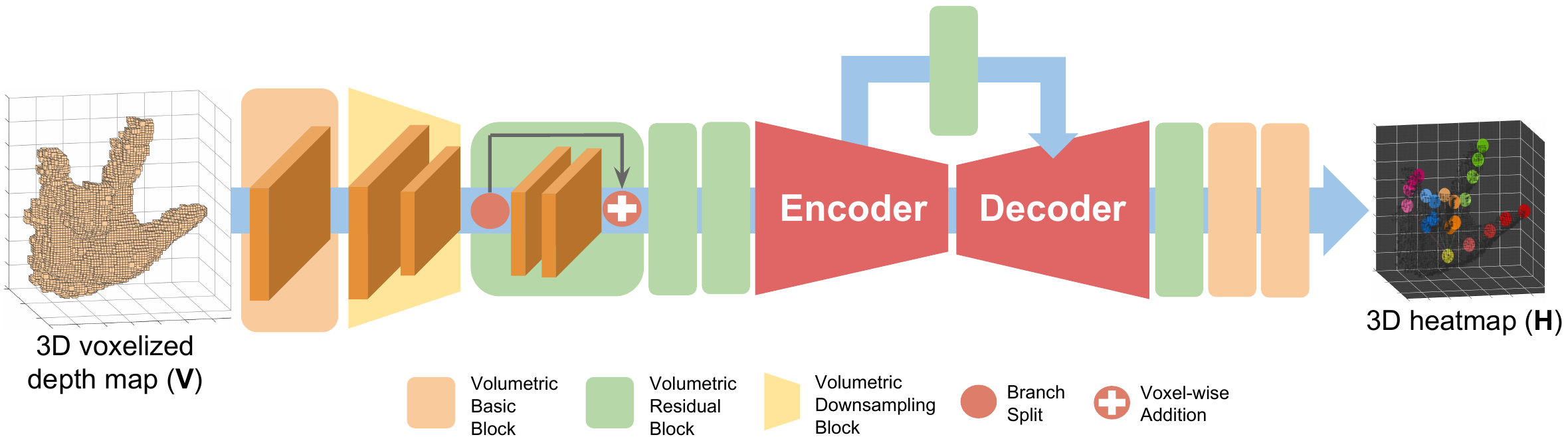}
\end{center}
\vspace*{-5mm}
   \caption{Overall architecture of the V2V-PoseNet. V2V-PoseNet takes voxelized input and estimates the per-voxel likelihood for each keypoint through encoder and decoder. To simplify the figure, we plotted each feature map without Z-axis and combined the 3D heatmaps of all keypoints in a single volume. Each color in the 3D heatmap indicates keypoints in the same finger.}
\vspace*{-3mm}
\label{fig:model_architecture}
\end{figure*}

{\bf Volumetric representation using depth information.}
Wu \etal~\cite{wu20153d} introduced the volumetric representation of a depth image and surpassed the existing hand-crafted descriptor-based methods in 3D shape classification and retrieval. They represented each voxel as a binary random variable and used a convolutional deep belief network to learn the probability distribution for each voxel. Several recent works~\cite{maturana2015voxnet,song2016deep} also represented 3D input data as a volumetric form for 3D object classification and detection. Our work follows the strategy from ~\cite{maturana2015voxnet}, wherein several types of volumetric representation (i.e., occupancy grid models) were proposed to fully utilize the rich source of 3D information and efficiently deal with large amounts of point cloud data. Their proposed CNN architecture and occupancy grids outperform those of Wu \etal~\cite{wu20153d}.

{\bf Input and output representation in 3D pose estimation.}
Most of the existing methods for 3D pose estimation from a single depth map~\cite{oberweger2015hands,oberweger2015training,bouchacourt2016disco,Wan_2017_CVPR,guo2017ren,guo2017towards,Oberweger_2017_ICCV_Workshops,chen2017pose,madadi2017end,fourure2017multi,haque2016towards} are based on the model in Figure~\ref{fig:comparison_io_type}(a) that takes a 2D depth image and directly regresses 3D coordinates. Recently, Ge \etal~\cite{ge20173d} and Deng \etal~\cite{deng2017hand3d} converted a 2D depth image to a truncated signed distance function-based 3D volumetric form and directly regressed 3D coordinates as shown in Figure~\ref{fig:comparison_io_type}(c). In 3D human pose estimation from a RGB image, Pavlakos \etal~\cite{pavlakos2016coarse} estimated the per-voxel likelihood for each body keypoint via 2D CNN as in the Figure~\ref{fig:comparison_io_type}(b). To estimate the per-voxel likelihood from an RGB image, they treated the discretized depth value as a channel of the feature map, which resulted in different kernels for each depth value. In contrast to all the above approaches, our proposed system estimates the per-voxel likelihood of each keypoint via the 3D fully convolutional network from the voxelized input as in Figure~\ref{fig:comparison_io_type}(d). To the best of our knowledge, our network is the first model to generate voxelized output from voxelized input using 3D CNN for 3D pose estimation.

\section{Overview of the proposed model}

The goal of our model is to estimate the 3D coordinates of all keypoints. First, we convert 2D depth images to 3D volumetric forms by reprojecting the points in the 3D space and discretizing the continuous space. After voxelizing the 2D depth image, the V2V-PoseNet takes the 3D voxelized data as an input and estimates the per-voxel likelihood for each keypoint. The position of the highest likelihood response for each keypoint is identified and warped to the real world coordinate, which becomes the final result of our model. Figure~\ref{fig:model_architecture} shows the overall architecture of the proposed V2V-PoseNet. We now describe the target object localization refinement strategy, the process of generating the input of the proposed model, V2V-PoseNet, and some related issues of the proposed approach in the following sections.

\begin{figure}[t]
\begin{center}
   \includegraphics[width=1.0\linewidth]{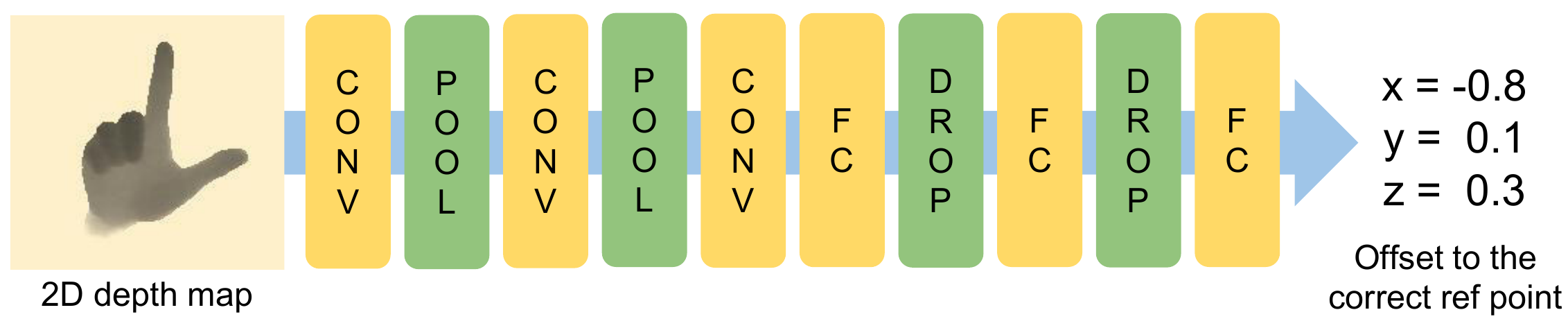}
\end{center}
\vspace*{-5mm}
   \caption{Reference point refining network. This network takes cropped depth image and outputs the 3D offset from the current reference point to the center of ground-truth joint locations.}
\vspace*{-3mm}
\label{fig:ref_refine_net}
\end{figure}

\section{Refining target object localization}
\label{refineRefPoint_Section}

To localize keypoints, such as hand or human body joints, a cubic box that contains the hand or human body in 3D space is a prerequisite. This cubic box is usually placed around the reference point, which is obtained using ground-truth joint position~\cite{oberweger2015hands,oberweger2015training,zhou2016model} or the center-of-mass after simple depth thresholding around the hand region~\cite{guo2017ren,guo2017towards,chen2017pose}. However, utilizing the ground-truth joint position is infeasible in real-world applications. Also, in general, using the center-of-mass calculated by simple depth thresholding does not guarantee that the object is correctly contained in the acquired cubic box due to the error in the center-of-mass calculations in cluttered scenes. For example, if other objects are near the target object, then the simple depth thresholding method cannot properly filter the other objects because it applies the same threshold value to all input data. Hence, the computed center-of-mass becomes erroneous, which results in a cubic box that contains only some part of the target object. To overcome these limitations, we train a simple 2D CNN following Oberweger \etal~\cite{Oberweger_2017_ICCV_Workshops} to obtain an accurate reference point as shown in Figure~\ref{fig:ref_refine_net}. This network takes a depth image, whose reference point is calculated by the simple depth thresholding around the hand region, and outputs 3D offset from the calculated reference point to the center of ground-truth joint locations. The refined reference point can be obtained by adding the output offset value of the network to the calculated reference point.

\section{Generating input of the proposed system}

To create the input of the proposed system, the 2D depth map should be converted to voxelized form. To voxelize the 2D depth map, we first reproject each pixel of the depth map to the 3D space. After reprojecting all depth pixels, the 3D space is discretized based on the pre-defined voxel size. Then, the target object is extracted by drawing the cubic box around the reference point obtained in Section~\ref{refineRefPoint_Section}. We set the voxel value of the network's input $V(i,j,k)$ as 1 if the voxel is occupied by any depth point and 0 otherwise.

\section{V2V-PoseNet}
\label{V2V-PoseNet_Section}

\subsection{Building block design}
We use four kinds of building blocks in designing the V2V-PoseNet. The first one is the \emph{volumetric basic block} that consists of a volumetric convolution, volumetric batch normalization~\cite{ioffe2015batch}, and the activation function (i.e., ReLU). This block is located in the first and last parts of the network. The second one is the \emph{volumetric residual block} extended from the 2D residual block of option B in ~\cite{he2016deep}. The third one is the \emph{volumetric downsampling block} that is identical to a volumetric max pooling layer. The last one is the \emph{volumetric upsampling block}, which consists of a volumetric deconvolution layer, volumetric batch normalization layer, and the activation function (i.e., ReLU). Adding the batch normalization layer and the activation function to the deconvolution layer helps to ease the learning procedure. The kernel size of the residual blocks is 3$\times$3$\times$3 and that of the downsampling and upsampling layers is 2$\times$2$\times$2 with stride 2.

\begin{figure}[t]
\begin{center}
   \includegraphics[width=1.0\linewidth]{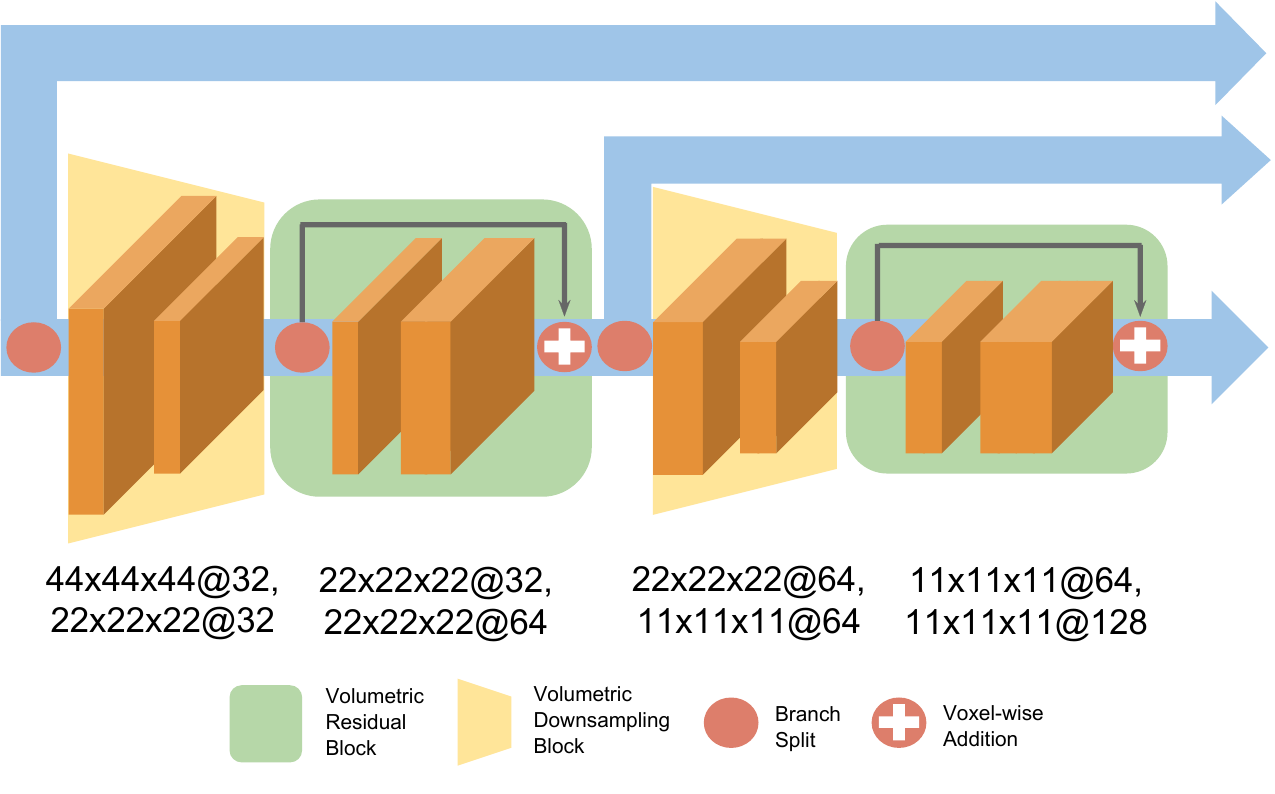}
\end{center}
\vspace*{-5mm}
   \caption{Encoder of the V2V-PoseNet. The numbers below each block indicate the spatial size and number of channels of each feature map. We plotted each feature map without Z-axis to simplify the figure.}
\vspace*{-4mm}
\label{fig:encoder}
\end{figure}

\subsection{Network design}
\begin{figure}[t]
\begin{center}
   \includegraphics[width=1.0\linewidth]{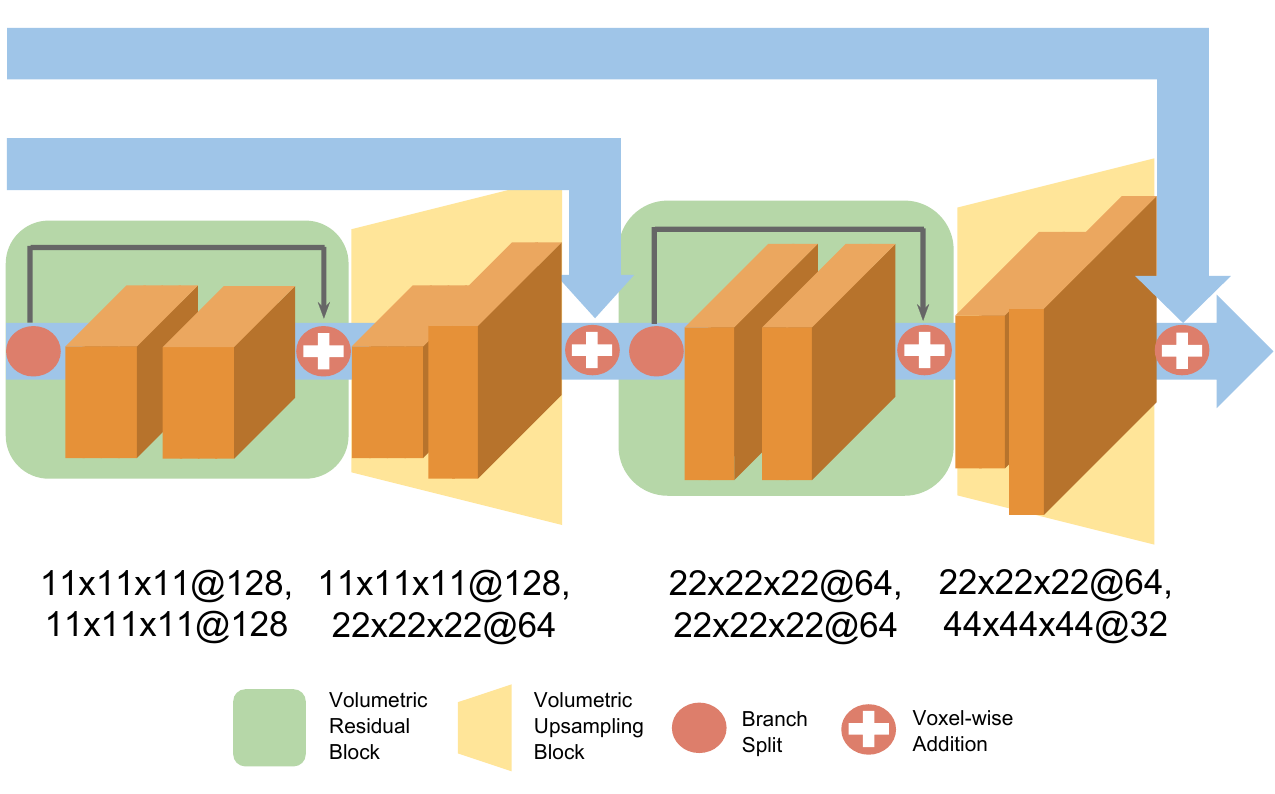}
\end{center}
\vspace*{-5mm}
   \caption{Decoder of the V2V-PoseNet. The numbers below each block indicate the spatial size and number of channels of each feature map. We plotted feature map without Z-axis to simplify the figure.}
\vspace*{-4mm}
\label{fig:decoder}
\end{figure}

The V2V-PoseNet performs voxel-to-voxel prediction. Thus, it is based on the 3D CNN architecture that treats the Z-axis as an additional spatial axis so that the kernel shape is $w$$\times$$h$$\times$$d$. Our network architecture is based on the hourglass model~\cite{newell2016stacked}, which was slightly modified for more accurate estimation. As the Figure~\ref{fig:model_architecture} shows, the network starts from the 7$\times$7$\times$7 volumetric basic block and the volumetric downsampling block. After downsampling the feature map, three consecutive residual blocks extract useful local features. The output of the residual blocks goes through the encoder and decoder described in Figures~\ref{fig:encoder} and ~\ref{fig:decoder}, respectively.

In the encoder, the volumetric downsampling block reduces the spatial size of the feature map while the volumetric residual bock increases the number of channels. It is empirically confirmed that this increase in the number of channels helps improve the performance in our experiments. On the other hand, in the decoder, the volumetric upsampling block enlarges the spatial size of the feature map. When upsampling, the network decreases the number of channels to compress the extracted features. The enlargement of the volumetric size in the decoder helps the network to densely localize keypoints because it reduces the stride between voxels in the feature map. The encoder and decoder are connected with the voxel-wise addition for each scale so that the decoder can upsample the feature map more stably. After passing the input through the encoder and decoder, the network predicts the per-voxel likelihood for each keypoint through two 1$\times$1$\times$1 volumetric basic blocks and one 1$\times$1$\times$1 volumetric convolutional layer.

\subsection{Network training}
To supervise the per-voxel likelihood for each keypoint, we generate 3D heatmap, wherein the mean of Gaussian peak is positioned at the ground-truth joint location as follows: 

\begin{equation}
H_{\mathrm{n}}^*(i,j,k) = \exp\left(-\frac{(i-i_{\mathrm{n}})^2+(j-j_{\mathrm{n}})^2+(k-k_{\mathrm{n}})^2}{2\sigma^2}\right),
\end{equation}
where $H_{n}^{*}$ is the ground-truth 3D heatmap of $n$th keypoint, ($i_n$,$j_n$,$k_n$) is the ground-truth voxel coordinate of $n$th keypoint, and $\sigma$ = 1.7 is the standard deviation of the Gaussian peak.

Also, we adopt the mean square error as a loss function $L$ as follows:

\begin{equation}
L = \sum_{n=1}^{N} \sum_{i,j,k} \|H_{\mathrm{n}}^*(i,j,k)-H_{\mathrm{n}}(i,j,k)\|^2,
\end{equation}
where $H_{n}^{*}$ and $H_{n}$ are the ground-truth and estimated heatmaps for $n$th keypoint, respectively, and $N$ denotes the number of keypoints.

\section{Implementation details}
The proposed V2V-PoseNet is trained in an end-to-end manner from scratch. All weights are initialized from the zero-mean Gaussian distribution with $\sigma$ = 0.001. Gradient vectors are calculated from the loss function and the weights are updated by the RMSProp~\cite{tieleman2012lecture} with a mini-batch size of 8. The learning rate is set to 2.5$\times$$10^{-4}$. The size of the input to the proposed system is 88$\times$88$\times$88. We perform data augmentation including rotation ([-40, 40] degrees in XY space), scaling ([0.8, 1.2] in 3D space), and translation ([-8, 8] voxels in 3D space). Our model is implemented by Torch7~\cite{collobert2011torch7} and the NVIDIA Titan X GPU is used for training and testing. We trained our model for 10 epochs.

\section{Experiment}

\subsection{Datasets}
{\bf ICVL Hand Posture Dataset.} 
The ICVL dataset~\cite{tang2014latent} consists of 330K training and 1.6K testing depth images. The frames are collected from 10 different subjects using Intel's Creative Interactive Gesture Camera~\cite{melax2013dynamics}. The annotation of hand pose contains 16 joints, which include three joints for each finger and one joint for the palm.

{\bf NYU Hand Pose Dataset.}
The NYU dataset~\cite{tompson2014real} consists of 72K training and 8.2K testing depth images. The training set is collected from subject A, whereas the testing set is collected from subjects A and B by three Kinects from different views. The annotations of hand pose contain 36 joints. Most of the previous works only used frames from the frontal view and 14 out of 36 joints in the evaluation, and we also followed them.

{\bf MSRA Hand Pose Dataset.}
The MSRA dataset~\cite{sun2015cascaded} contains 9 subjects with 17 gestures for each subject. Intel's Creative Interactive Gesture Camera~\cite{melax2013dynamics} captured 76K depth images with 21 annotated joints. For evaluation, the leave-one-subject-out cross-validation strategy is utilized. 

{\bf HANDS 2017 Frame-based 3D Hand Pose Estimation Challenge Dataset.}
The HANDS 2017 frame-based 3D hand pose estimation challenge dataset~\cite{yuan20172017} consists of 957K training and 295K testing depth images that are sampled from BigHand2.2M~\cite{Yuan_2017_CVPR} and First-Person Hand Action~\cite{garcia2017first} datasets. There are five subjects in the training set and ten subjects in the testing stage, including five unseen subjects. The ground-truth of this dataset is the 3D coordinates of 21 hand joints.

{\bf ITOP Human Pose Dataset.}
The ITOP dataset~\cite{haque2016towards} consists of 40K training and 10K testing depth images for each of the front-view and top-view tracks. This dataset contains depth images with 20 actors who perform 15 sequences each and is recorded by two Asus Xtion Pro cameras. The ground-truth of this dataset is the 3D coordinates of 15 body joints.

\subsection{Evaluation metrics}

We used 3D distance error and percentage of success frame metrics for 3D hand pose estimation following~\cite{tang2014latent,sun2015cascaded}. For 3D human pose estimation, we used mean average precision (mAP) that is defined as the detected ratio of all human body joints based on 10 cm rule following~\cite{haque2016towards,yub2015random}.

\subsection{Ablation study}
We used NYU hand pose dataset~\cite{tompson2014real} to analyze each component of our model because this dataset is challenging and far from saturated.

\begin{table}[]
\centering
\setlength\tabcolsep{1.0pt}
\def\arraystretch{1.1}
\begin{tabular}{L{2.6cm}|C{2.5cm}C{3cm}}
\specialrule{.1em}{.05em}{.05em} 
   Input \textbackslash Output & 3D Coordinates & Per-voxel likelihood  \\ \hline
2D depth map & 18.85 (21.1 M) & 13.01 (4.6 M) \\ 
3D voxelized grid & 16.78 (457.5 M)  & \textbf{10.37 (3.4 M)} \\ \specialrule{.1em}{.05em}{.05em} 
\end{tabular}
\vspace*{-3mm}
\caption{Average 3D distance error (mm) and number of parameter comparison of the input and output types in the NYU dataset. The number in the parenthesis denotes the number of parameters. The visualized model for each input and output type is shown in Figure~\ref{fig:comparison_io_type}.}
\label{table:nyu_io_type}
\end{table}

\begin{table}
\centering
\setlength\tabcolsep{1.0pt}
\def\arraystretch{1.1}
\begin{tabular}{L{3.7cm}C{3.7cm}}
\specialrule{.1em}{.05em}{.05em} 
   Methods  &  Average 3D distance error \\ \hline
Baseline      & 11.14 mm \\ 
 + Localization refinement      &  9.22 mm \\ 
 + Epoch ensemble      & 8.42 mm \\ \specialrule{.1em}{.05em}{.05em}
\end{tabular}
\vspace*{-3mm}
\caption{Effect of localization refinement and epoch ensemble. The average 3D distance error is calculated in the NYU dataset.}
\vspace*{-4mm}
\label{table:nyu_R_E}
\end{table}

\begin{figure*}
\begin{center}
\includegraphics[width=1.0\linewidth]{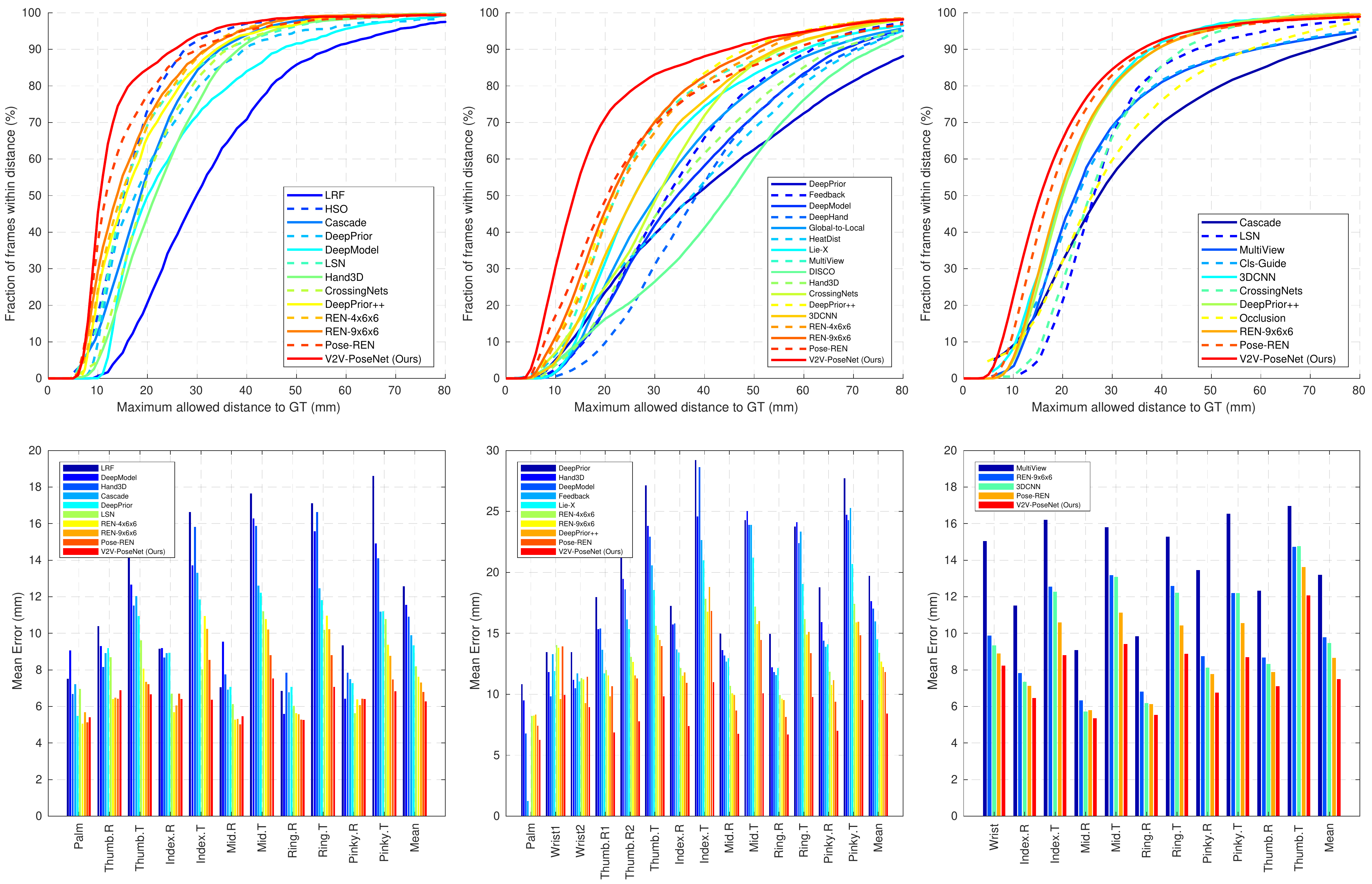}
\end{center}
\vspace*{-5mm}
   \caption{Comparison of the proposed method (V2V-PoseNet) with state-of-the-art methods. Top row: the percentage of success frames over different error thresholds. Bottom row: 3D distance errors per hand keypoints. Left: ICVL dataset, middle: NYU dataset, right: MSRA dataset.}
\label{fig:comparison_with_stoa}
\end{figure*}

\begin{figure}[t]
\begin{center}
   \includegraphics[width=1.0\linewidth]{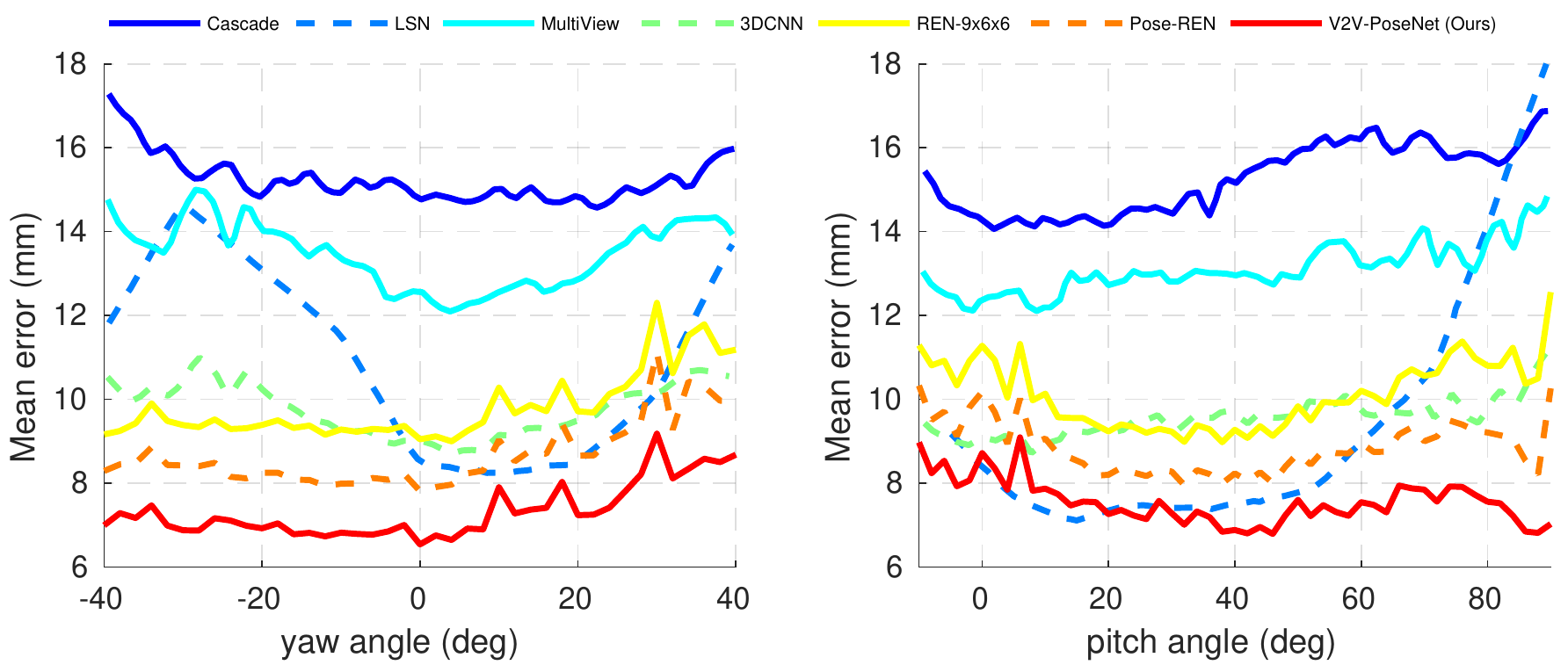}
\end{center}
\vspace*{-5mm}
   \caption{Comparison of average 3D distance error over different yaw (left) and pitch (right) angles on the MSRA dataset.}
\vspace*{-3mm}
\label{fig:msra_yaw_pitch}
\end{figure}

\begin{table*}[t]
\centering
\setlength\tabcolsep{1.0pt}
\def\arraystretch{1.1}
\begin{subtable}[t]{.34\textwidth}
\centering
\begin{tabular}[t]{C{3.1cm}C{2.5cm}}
\specialrule{.1em}{.05em}{.05em} 
   Methods  &  Mean error (mm)  \\ \hline
LRF     & 12.58  \\ 
DeepModel     &  11.56 \\ 
Hand3D     &  10.9  \\ 
CDO & 10.5 \\
DeepPrior      & 10.4  \\ 
CrossingNets    &  10.2  \\ 
Cascade      & 9.9  \\
JTSC    &  9.16  \\ 
DeepPrior++      &  8.1  \\ 
REN-4x6x6	& 7.63 \\
REN-9x6x6      &  7.31  \\ 
Pose-REN     &  6.79  \\
\textbf{V2V-PoseNet (Ours)}      &  \textbf{6.28}  \\ \specialrule{.1em}{.05em}{.05em} 
\end{tabular}
\caption{ICVL}
\end{subtable}%
\begin{subtable}[t]{.34\textwidth}
\centering
\begin{tabular}[t]{C{3.1cm}C{2.5cm}}
\specialrule{.1em}{.05em}{.05em} 
   Methods  &  Mean error (mm)  \\ \hline
DISCO & 20.7\\
DeepPrior     & 19.73  \\ 
Hand3D     &  17.6 \\ 
DeepModel    &  17.04  \\ 
JTSC    &  16.8  \\ 
Feedback     &  15.97  \\ 
Global-to-Local     &  15.60  \\ 
Lie-X      &  14.51  \\ 
3DCNN     &  14.1  \\ 
REN-4x6x6	& 13.39 \\
REN-9x6x6    &  12.69  \\ 
DeepPrior++      &  12.24  \\ 
Pose-REN     &  11.81  \\
\textbf{V2V-PoseNet (Ours)}      &  \textbf{8.42}  \\ \specialrule{.1em}{.05em}{.05em} 
\end{tabular}
\caption{NYU}
\end{subtable}%
\begin{subtable}[t]{.34\textwidth}
\centering
\begin{tabular}[t]{C{3.1cm}C{2.5cm}}
\specialrule{.1em}{.05em}{.05em} 
   Methods  &  Mean error (mm)  \\ \hline
Cascade     & 15.2  \\
Cls-Guide& 13.7 \\
MultiView    & 13.2  \\
Occlusion    & 12.8  \\
CrossingNets     &  12.2  \\  
REN-9x6x6     &  9.7  \\ 
DeepPrior++      &  9.5 \\ 
Pose-REN     &  8.65  \\
\textbf{V2V-PoseNet (Ours)}      &  \textbf{7.49}  \\ \specialrule{.1em}{.05em}{.05em} 
\end{tabular}
\caption{MSRA}
\end{subtable}%
\caption{Comparison of the proposed method (V2V-PoseNet) with state-of-the-art methods on the three 3D hand pose datasets. Mean error indicates the average 3D distance error.}
\vspace*{-3mm}
\label{table:comparison_with_stoa}
\end{table*}

{\bf 3D representation and per-voxel likelihood estimation.}
To demonstrate the validity of the 3D representation of the input and per-voxel likelihood estimation, we compared the performances of the four different combinations of the input and output forms in Table~\ref{table:nyu_io_type}. As the table shows, converting the input representation type from the 2D depth map to 3D voxelized form (also converting the model from 2D CNN to 3D CNN) substantially improves performance, regardless of output representation. This justifies the effectiveness of the proposed 3D input representation that is free from perspective distortion. The results also show that converting the output representation from the 3D coordinates to the per-voxel likelihood increases the performance significantly, regardless of the input type. Among the four combinations, \emph{voxel-to-voxel} gives the best performance even with the smallest number of parameters. Hence, the superiority of the \emph{voxel-to-voxel} prediction scheme compared with other input and output combinations is clearly justified.

To fairly compare four combinations, we used the same network building blocks and design, which were introduced in Section~\ref{V2V-PoseNet_Section}. The only difference is that the model for the per-voxel likelihood estimation is fully convolutional, whereas for the coordinate regression, we used fully connected layers at the end of the network. Simply converting \emph{voxel-to-voxel} to \emph{pixel-to-voxel} decreases the number of parameters because the model is changed from the 3D CNN to the 2D CNN. To compensate for this change, we doubled the number of channels of each feature map in the \emph{pixel-to-voxel} model. If the number of channels is not doubled, then the performance was degraded. For all four models, we used 48$\times$48 depth map or 48$\times$48$\times$48 voxelized grid as input because the original size (88$\times$88$\times$88) does not fit into GPU memory in the case of \emph{voxel-to-coordinates}.

{\bf Refining localization of the target object.}
To demonstrate the importance of the localization refining procedure in Section~\ref{refineRefPoint_Section}, we compared the performance of two with and without the localization refinement step. As shown in Table~\ref{table:nyu_R_E}, the refined reference points significantly boost the accuracy of our model, which shows that the reference point refining procedure has a crucial influence on the performance.

{\bf Epoch ensemble.}
To obtain more accurate and robust estimation, we applied a simple ensemble technique that we call \emph{epoch ensemble}. The epoch ensemble averages the estimations from several epochs. Specifically, we save the trained model for each epoch in the training stage and then in the testing stage, we average all the estimated 3D coordinates from the trained models. As we trained our model by 10 epochs, we used 10 models to obtain the final estimation. Epoch ensemble has no influence in running time when each model is running in different GPUs. However, in a single-GPU environment, epoch ensemble linearly increases running time. The effect of epoch ensemble is shown in Table~\ref{table:nyu_R_E}.

\subsection{Comparison with state-of-the-art methods}
We compared the performance of the V2V-PoseNet on the three 3D hand pose estimation datasets (ICVL~\cite{tang2014latent}, NYU~\cite{tompson2014real}, and MSRA~\cite{sun2015cascaded}) with most of the state-of-the-art methods, which include latent random forest (LRF)~\cite{tang2014latent}, cascaded hand pose regression (Cascade)~\cite{sun2015cascaded}, DeepPrior with refinement (DeepPrior)~\cite{oberweger2015hands}, feedback loop training method (Feedback)~\cite{oberweger2015training}, hand model based method (DeepModel)~\cite{zhou2016model}, hierarchical sampling optimization (HSO)~\cite{tang2015opening}, local surface normals (LSN)~\cite{wan2016hand}, multi-view CNN (MultiView)~\cite{ge2016robust}, DISCO~\cite{bouchacourt2016disco}, Hand3D~\cite{deng2017hand3d}, DeepHand~\cite{sinha2016deephand}, lie-x group based method (Lie-X)~\cite{xu2017lie}, improved DeepPrior (DeepPrior++)~\cite{Oberweger_2017_ICCV_Workshops}, region ensemble network (REN-4$\times$6$\times$6~\cite{guo2017ren}, REN-9$\times$6$\times$6~\cite{guo2017towards}), CrossingNets~\cite{Wan_2017_CVPR}, pose-guided REN (Pose-REN)~\cite{chen2017pose}, global-to-local prediction method (Global-to-Local)~\cite{madadi2017end}, classification-guided approach (Cls-Guide)~\cite{yang2016hand}, 3DCNN based method (3DCNN)~\cite{ge20173d},  occlusion aware based method (Occlusion)~\cite{madadi2017occlusion}, and hallucinating heat distribution method (HeatDist)~\cite{Choi_2017_ICCV}. Some reported results of previous works~\cite{tang2014latent,oberweger2015hands,oberweger2015training,zhou2016model,Oberweger_2017_ICCV_Workshops,guo2017ren,guo2017towards,chen2017pose,xu2017lie} are calculated by prediction labels available online. Other results~\cite{sun2015cascaded,tang2015opening,wan2016hand,sinha2016deephand,ge2016robust,deng2017hand3d,Wan_2017_CVPR,madadi2017end,ge20173d,madadi2017occlusion,Choi_2017_ICCV,bouchacourt2016disco,yang2016hand} are calculated from the figures and tables of their papers.

\begin{table}
\centering
\setlength\tabcolsep{1.0pt}
\def\arraystretch{1.1}
\begin{tabular}{L{4.5cm}C{3.7cm}}
\specialrule{.1em}{.05em}{.05em} 
   Team name &  Average 3D distance error \\ \hline
V2V-PoseNet (Ours)      & \textbf{9.95 mm} \\ 
NVResearch and UMontreal      &  10.18 mm \\ 
NTU     & 11.30 mm \\ 
THU VCLab      & 11.70 mm \\ 
NAIST RVLab      & 11.90 mm \\ \specialrule{.1em}{.05em}{.05em}
\end{tabular}
\vspace*{-3mm}
\caption{The top-5 results of the HANDS 2017 frame-based 3D hand pose estimation challenge.}
\vspace*{-5.5mm}
\label{table:hands2017_result}
\end{table}

\begin{table*}
\centering
\setlength\tabcolsep{1.0pt}
\def\arraystretch{1.1}
\begin{tabular}{L{1.5cm}|C{1.0cm}C{1.1cm}C{1.1cm}C{1.0cm}C{1.2cm}C{2.0cm}|C{1.0cm}C{1.1cm}C{1.1cm}C{1.0cm}C{1.2cm}C{2.0cm}}
\specialrule{.1em}{.05em}{.05em} 
 & \multicolumn{6}{c|}{mAP (front-view)}  &  \multicolumn{6}{c}{mAP (top-view)} \\ \hline
   Body part & RF & RTW & IEF &VI & REN-9x6x6 & V2V-PoseNet (Ours)  & RF & RTW & IEF & VI & REN-9x6x6 & V2V-PoseNet (Ours) \\ \hline
Head & 63.8 & 97.8 & 96.2 & 98.1 & \textbf{98.7} & 98.29 & 95.4 & \textbf{98.4} & 83.8 & 98.1 & 98.2 & \textbf{98.4} \\ 
Neck & 86.4 & 95.8 & 85.2 & 97.5 & \textbf{99.4} & 99.07 & 98.5 & 82.2 & 50.0 & 97.6 & 98.9 & \textbf{98.91} \\ 
Shoulders & 83.3 & 94.1 & 77.2 & 96.5 & 96.1 & \textbf{97.18} & 89.0 & 91.8 & 67.3 & 96.1 & 96.6 & \textbf{96.87} \\ 
Elbows & 73.2 & 77.9 & 45.4 &  73.3 & 74.7 & \textbf{80.42} & 57.4 & 80.1 & 40.2 & \textbf{86.2} & 74.4 & 79.16 \\ 
Hands & 51.3 & \textbf{70.5} & 30.9 & 68.7 & 55.2 & 67.26 & 49.1 & 76.9 & 39.0 & \textbf{85.5} & 50.7 & 62.44\\ 
Torso & 65.0 & 93.8 & 84.7 & 85.6 & 98.7 & \textbf{98.73} & 80.5 & 68.2 & 30.5 & 72.9 & \textbf{98.1} & 97.78 \\ 
Hip & 50.8 & 80.3 & 83.5 & 72.0 & 91.8 & \textbf{93.23} & 20.0 & 55.7 & 38.9 & 61.2 & 85.5 & \textbf{86.91}\\ 
Knees & 65.7 & 68.8 & 81.8 & 69.0 & 89.0 & \textbf{91.80} & 2.6 & 53.9 & 54.0 & 51.6 & 70.0 & \textbf{83.28}\\ 
Feet & 61.3 & 68.4 & 80.9 & 60.8 & 81.1 & \textbf{87.6} & 0.0 & 28.7 & 62.4 & 51.5 & 41.6 & \textbf{69.62}\\ \hhline{-------------}
Mean & 65.8 & 80.5 & 71.0 & 77.4 & 84.9 & \textbf{88.74} & 47.4 & 68.2 & 51.2 & 75.5 & 75.5 & \textbf{83.44}\\ \specialrule{.1em}{.05em}{.05em} 
\end{tabular}
\vspace*{-3mm}
\caption{Comparison of the proposed method (V2V-PoseNet) with state-of-the-art methods on the front and top views of the ITOP dataset.}
\label{table:comparison_with_stoa_itop}
\end{table*}

As shown in Figure~\ref{fig:comparison_with_stoa} and Table~\ref{table:comparison_with_stoa}, our method outperforms all existing methods on the three 3D hand pose estimation datasets in standard evaluation metrics. This shows the superiority of \emph{voxel-to-voxel} prediction, which is firstly used in 3D hand pose estimation. The performance gap between ours and the previous works is largest on the NYU dataset that is very challenging and far from saturated. We additionally measured the average 3D distance error distribution over various yaw and pitch angles on the MSRA dataset following the protocol of previous works~\cite{sun2015cascaded} as in Figure~\ref{fig:msra_yaw_pitch}. As it demonstrates, our method provides superior results in almost all of yaw and pitch angles.

Our method also placed first in the HANDS 2017 frame-based 3D hand pose estimation challenge~\cite{yuan20172017}. The top-5 results comparisons are shown in Table~\ref{table:hands2017_result}. As shown in the table, the proposed V2V-PoseNet outperforms other participants. A more detailed analysis of the challenge results is covered in ~\cite{yuan20183d}.

We also evaluated the performance of the proposed system on the ITOP 3D human pose estimation dataset~\cite{haque2016towards}. We compared the system with state-of-the-art works, which include random forest-based method (RF)~\cite{shotton2013real}, RTW~\cite{yub2015random}, IEF~\cite{carreira2016human}, viewpoint-invariant feature-based method (VI)~\cite{haque2016towards}, and  REN-9x6x6~\cite{guo2017towards}. The score of each method is obtained from ~\cite{haque2016towards,guo2017towards}. As shown in Table~\ref{table:comparison_with_stoa_itop}, the proposed system outperforms all the existing methods by a large margin in both of views, which indicates that our model can be applied to not only 3D hand pose estimation, but also other challenging problems such as 3D human pose estimation from the front- and top-views. 

\begin{figure*}
\begin{center}
   \includegraphics[width=1.0\linewidth]{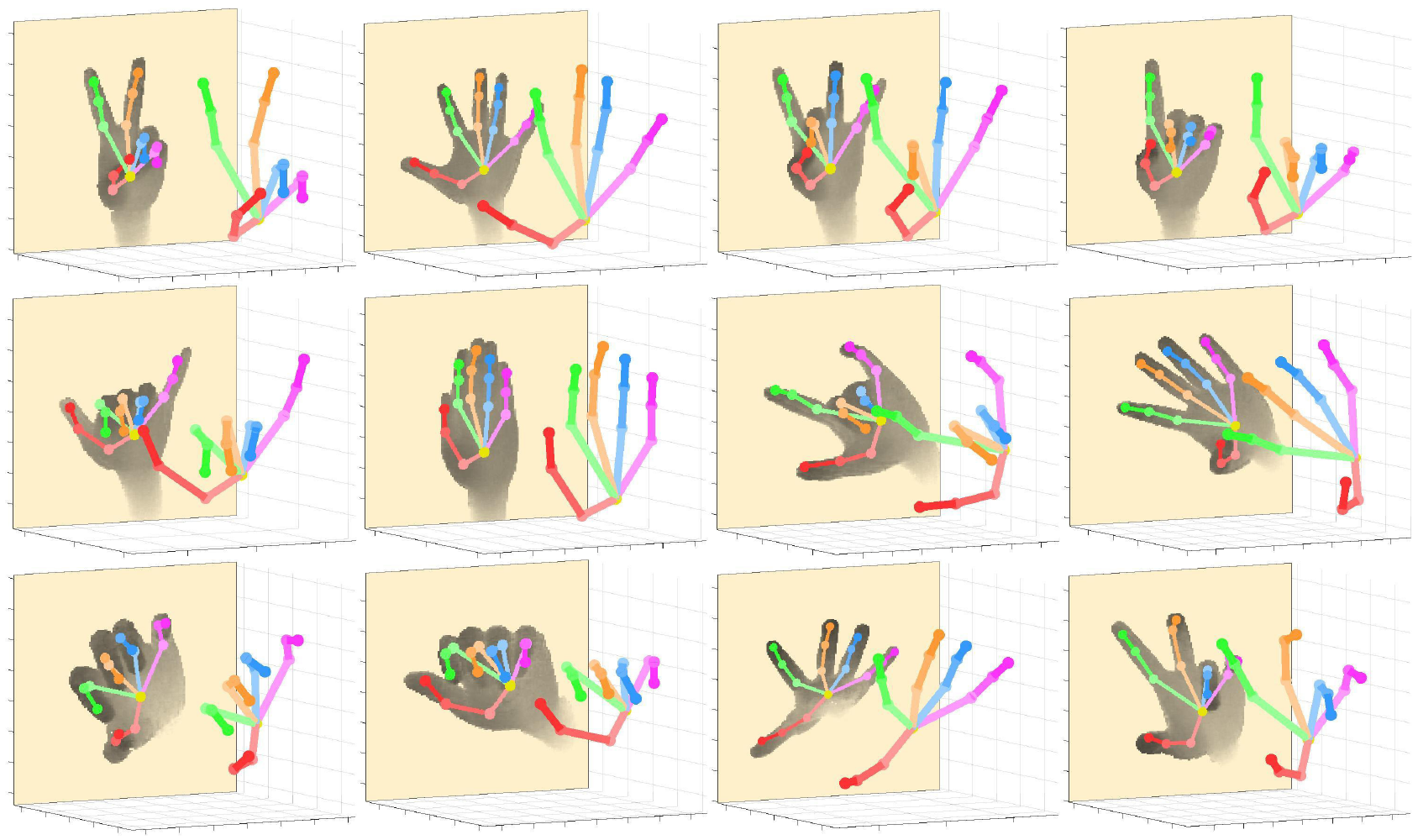}
\end{center}
\vspace*{-6mm}
   \caption{Qualitative results of our V2V-PoseNet on the ICVL dataset. Backgrounds are removed to make them visually pleasing.}
\vspace*{-3mm}
\label{fig:qualitative_icvl}
\end{figure*}

\begin{figure*}
\begin{center}
   \includegraphics[width=1.0\linewidth]{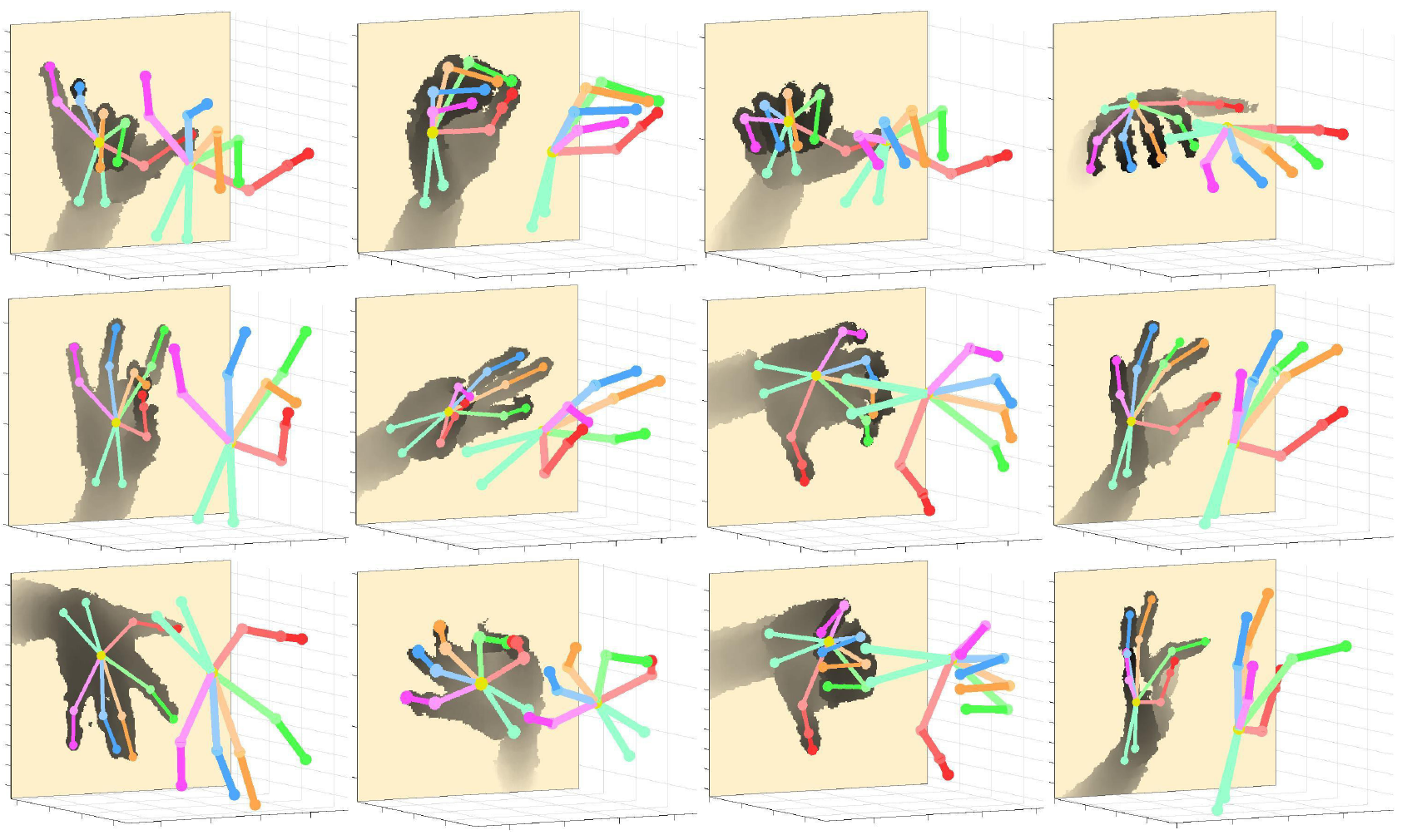}
\end{center}
\vspace*{-6mm}
   \caption{Qualitative results of our V2V-PoseNet on the NYU dataset. Backgrounds are removed to make them visually pleasing.}
\vspace*{-3mm}
\label{fig:qualitative_nyu}
\end{figure*}

\begin{figure*}
\begin{center}
   \includegraphics[width=1.0\linewidth]{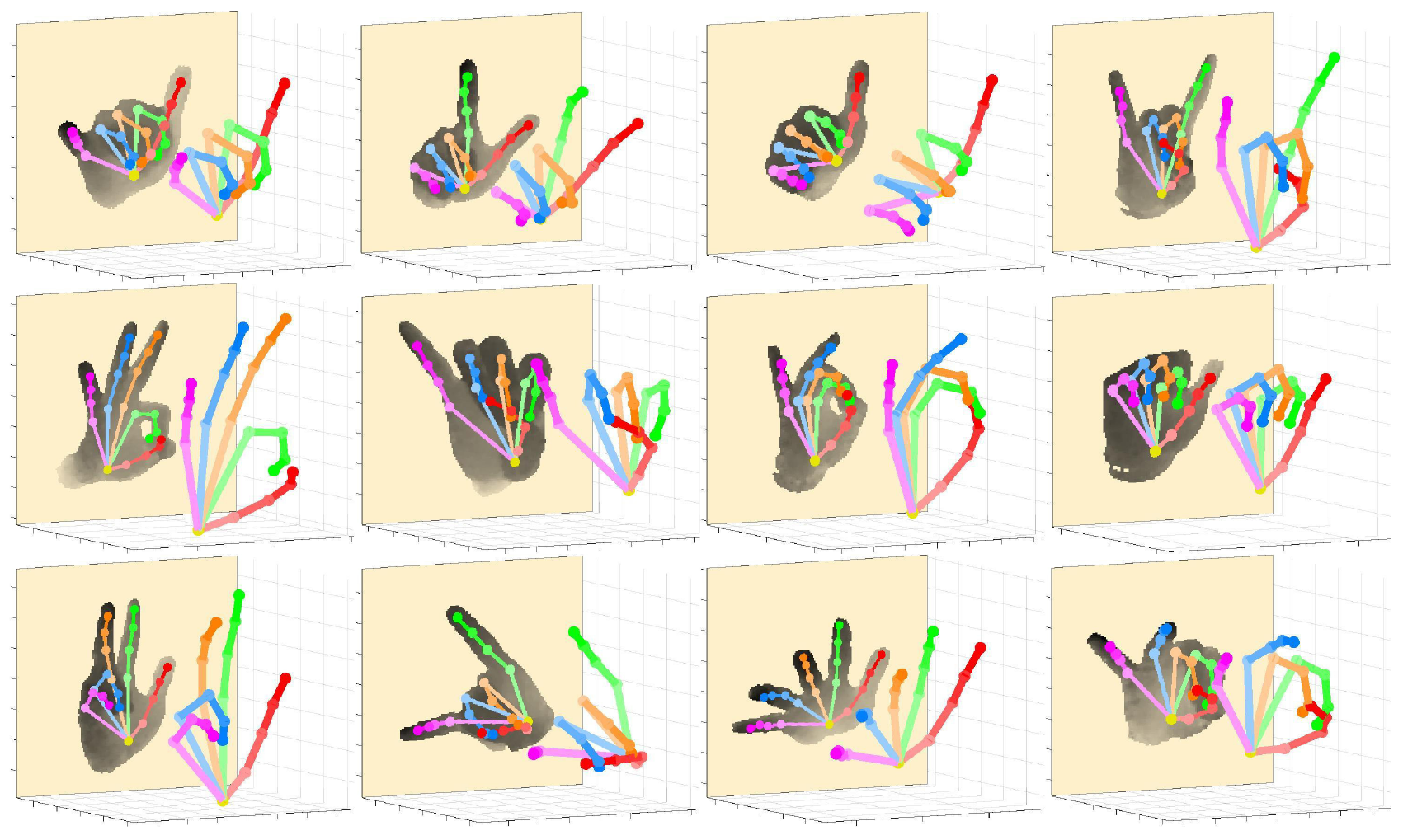}
\end{center}
\vspace*{-6mm}
   \caption{Qualitative results of our V2V-PoseNet on the MSRA dataset. Backgrounds are removed to make them visually pleasing.}
\vspace*{-3mm}
\label{fig:qualitative_msra}
\end{figure*}

\begin{figure*}
\begin{center}
   \includegraphics[width=1.0\linewidth]{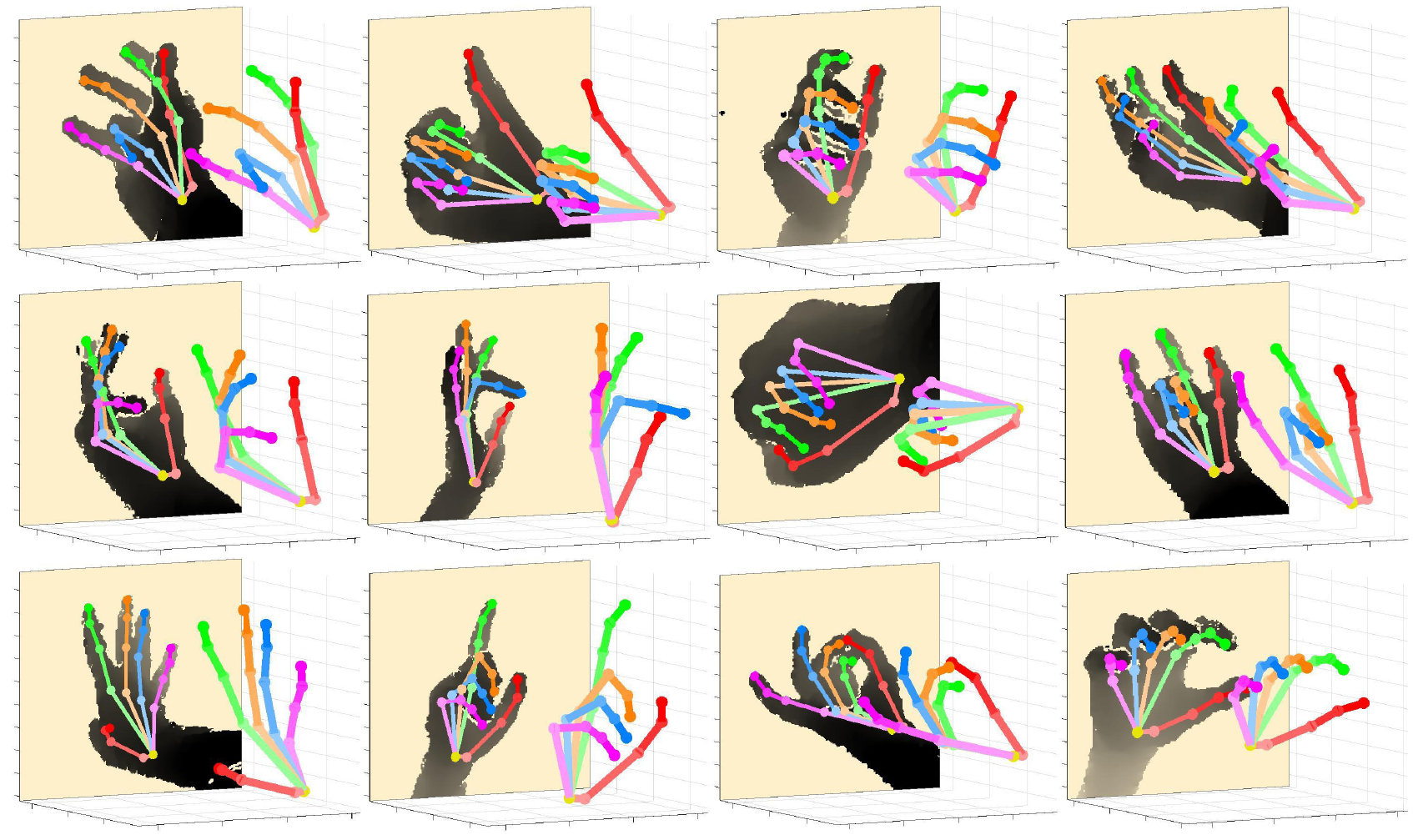}
\end{center}
\vspace*{-6mm}
   \caption{Qualitative results of our V2V-PoseNet on the HANDS 2017 frame-based 3D hand pose estimation challenge dataset. Backgrounds are removed to make them visually pleasing.}
\vspace*{-3mm}
\label{fig:qualitative_hands2017}
\end{figure*}

\begin{figure*}
\begin{center}
   \includegraphics[width=1.0\linewidth]{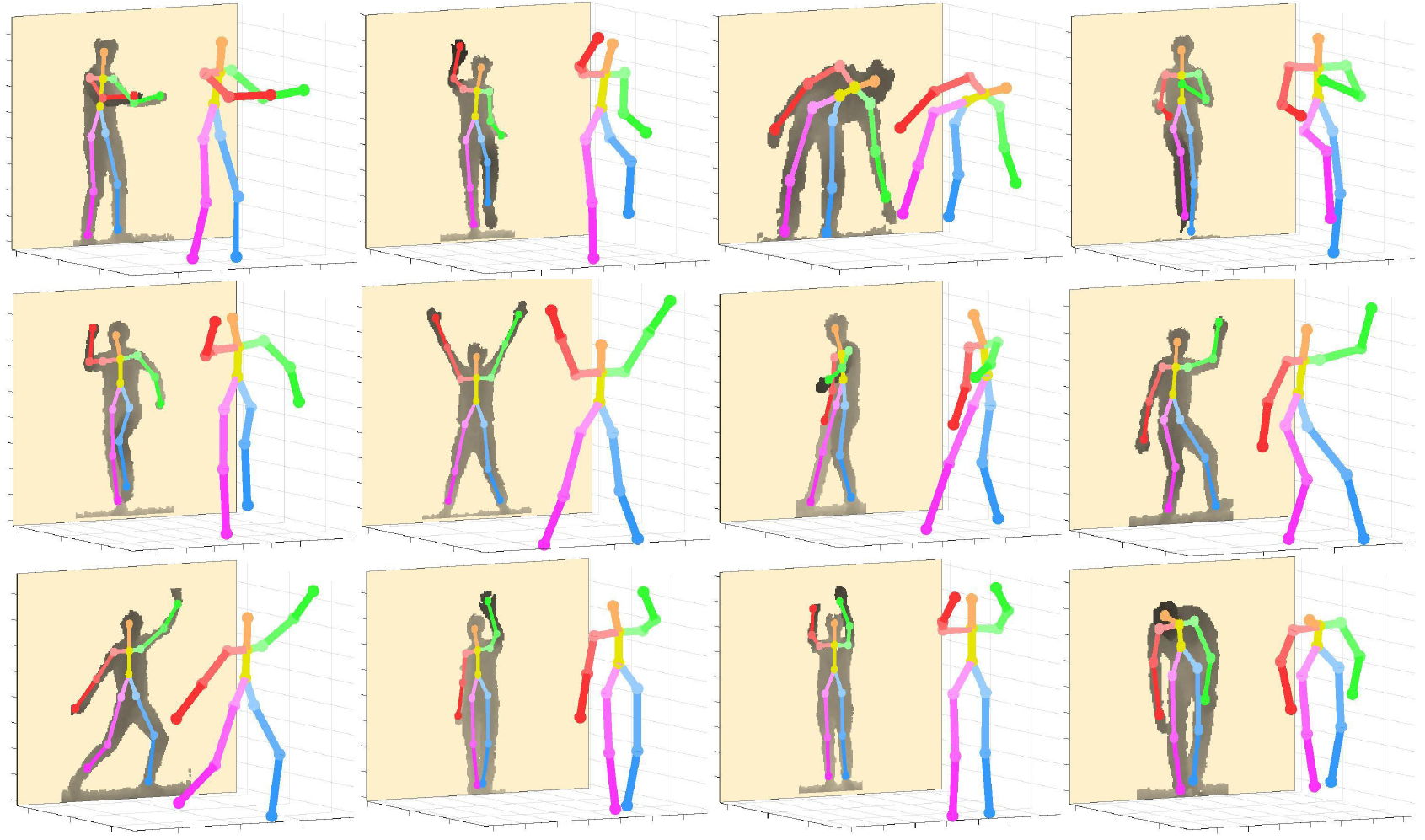}
\end{center}
\vspace*{-6mm}
   \caption{Qualitative results of our V2V-PoseNet on the ITOP dataset (front-view). Backgrounds are removed to make them visually pleasing.}
\vspace*{-3mm}
\label{fig:qualitative_itop_front}
\end{figure*}

\begin{figure*}
\begin{center}
   \includegraphics[width=1.0\linewidth]{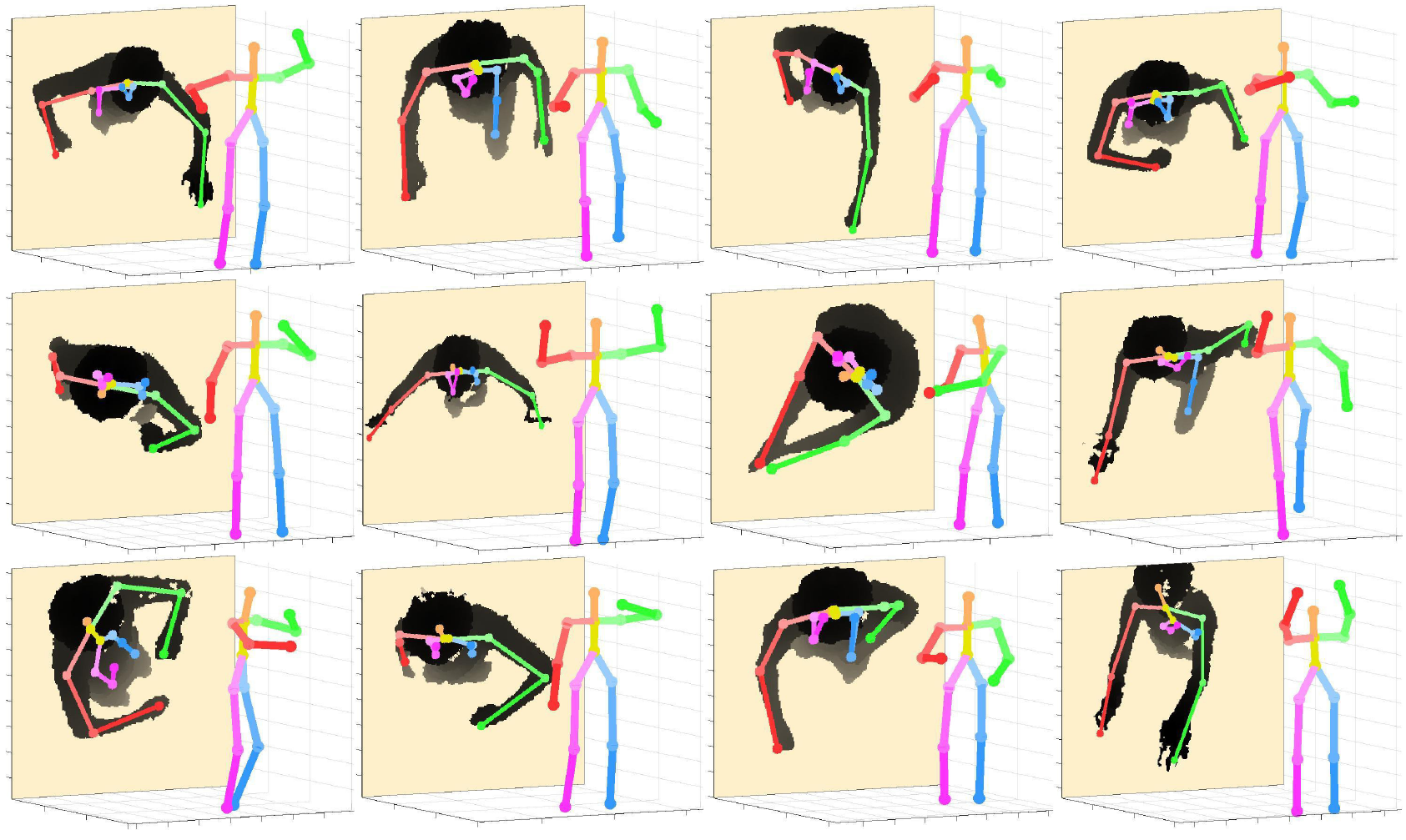}
\end{center}
\vspace*{-6mm}
   \caption{Qualitative results of our V2V-PoseNet on the ITOP dataset (top-view). Backgrounds are removed to make them visually pleasing.}
\vspace*{-3mm}
\label{fig:qualitative_itop_top}
\end{figure*}

The qualitative results of the V2V-PoseNet on the ICVL, NYU, MSRA, HANDS 2017, ITOP front-view, and ITOP top-view datasets are shown in Figure~\ref{fig:qualitative_icvl}, ~\ref{fig:qualitative_nyu}, ~\ref{fig:qualitative_msra}, ~\ref{fig:qualitative_hands2017}, ~\ref{fig:qualitative_itop_front}, and ~\ref{fig:qualitative_itop_top}, respectively.

\subsection{Computational complexity}
We investigated the computational complexity of the proposed method. The training time of the V2V-PoseNet is two days for ICVL dataset, 12 hours for NYU and MSRA datasets, six days for HANDS 2017 challenge dataset, and three hours for ITOP dataset. The testing time is 3.5 fps when using 10 models for epoch ensemble, but can accelerate to 35 fps in a multi-GPU environment, which shows the applicability of the proposed method to real-time applications. The most time-consuming step is the input generation that includes reference point refinement and voxelizing the depth map. This step takes 23 ms and most of the time is spent on voxelizing. The next step is network forwarding, which takes 5 ms and takes 0.5 ms to extract 3D coordinates from the 3D heatmap. Note that our model outperforms previous works by a large margin without epoch ensemble on the ICVL, NYU, MSRA, and ITOP datasets while running in real-time using a single GPU.

\section{Conclusion}
We proposed a novel and powerful network, V2V-PoseNet, for 3D hand and human pose estimation from a single depth map. To overcome the drawbacks of previous works, we converted 2D depth map into the 3D voxel representation and processed it using our 3D CNN model. Also, instead of directly regressing 3D coordinates of keypoints, we estimated the per-voxel likelihood for each keypoint. Those two conversions boost the performance significantly and make the proposed V2V-PoseNet outperform previous works on the three 3D hand and one 3D human pose estimation datasets by a large margin. It also allows us to win the 3D hand pose estimation challenge. As \emph{voxel-to-voxel} prediction is firstly tried in 3D hand and human pose estimation from a single depth map, we hope this work to provide a new way of accurate 3D pose estimation.

\clearpage
\clearpage

{\small
\bibliographystyle{ieee}
\bibliography{egbib}
}

\end{document}